# A Qualitative Comparative Evaluation of Cognitive and Generative Theories

**Paul S. Rosenbloom**  ROSENBLOOM@USC.EDU
Thomas Lord Department of Computer Science & Institute for Creative Technologies, University of Southern California, 230 S Thurston Ave, Los Angeles, CA 90049 USA

### Abstract

Evaluation is a critical activity associated with any theory. Yet this has proven to be an exceptionally challenging activity for theories based on cognitive architectures. For an overlapping set of reasons, evaluation can also be challenging for theories based on generative neural architectures. This dual challenge is approached here by leveraging a broad perspective on theory evaluation to yield a wide-ranging, albeit qualitative, comparison of whole-mind-oriented cognitive and generative architectures and the full systems that are based on these architectures.

## 1. Introduction

A theory, at least for our purposes here, can be considered generically as a body of material that is about some phenomena (Rosenbloom, 2026). This is intended to include both what are traditionally considered as theories and models, however the distinction between the two might be drawn. The theories of central concern here are ones that are geared towards modeling whole minds, whether human, artificial, or some abstraction over both. This thus includes examples not only from cognitive science but also from artificial intelligence (AI) and artificial general intelligence (AGI), whether symbolic, neural, or hybrid.

Such theories can typically be partitioned into *architectures* versus *variable content* (Rosenbloom, 2026). Architectures encapsulate the requisite fixed structures and processes, such as long-term and working memories, and mechanisms for learning and decision making. Variable content comprises the knowledge, skills, links, parameter values, etc. that is required on top of the architectures to yield effective behavior. Together, architectures and variable content yield *systems*.

In this work, two general classes of such theories are compared: (1) *cognitive theories* based on symbolic technologies and their hybrid variants (Kotseruba & Tsotsos, 2020, provides a compendium of many such theories); and (2) *generative theories* based on neural network technologies, most particularly *large language models* (e.g., Brown *et al*., 2020). These two classes were chosen for comparison because they represent, respectively, the leading longstanding approach to developing theories of whole minds and a highly significant recent challenger to this leadership.[1]

---

[1] One important class of theories not included in this analysis is *whole-brain* theories that are built around architectures, such as Leabra (O'Reilly & Munakata, 2000) and Spaun (Eliasmith, 2013), that while neural

When comparing theories of whole minds one of the key questions must be the criteria by which they are to be evaluated. Accuracy with respect to the phenomena is typically the preeminent criterion. However, it is both a very difficult criterion to apply to theories of whole minds and not by far the only criterion that matters. A recent attempt to synthesize across a number of existing definitions of "theory" plus prior work on evaluating cognitive architectures (Newell, 1990; Langley, Laird & Rogers, 2009; Kotseruba & Tsotsos, 2020; Lieto, 2021) has alternatively led to the identification of a set of criteria that is notable for its breadth, depth, and overall structure (Rosenbloom, 2026). As shown in Figure 1, it turns out to be most appropriately represented as a directed graph in which nodes represent criteria and arrows indicate dependencies among them.

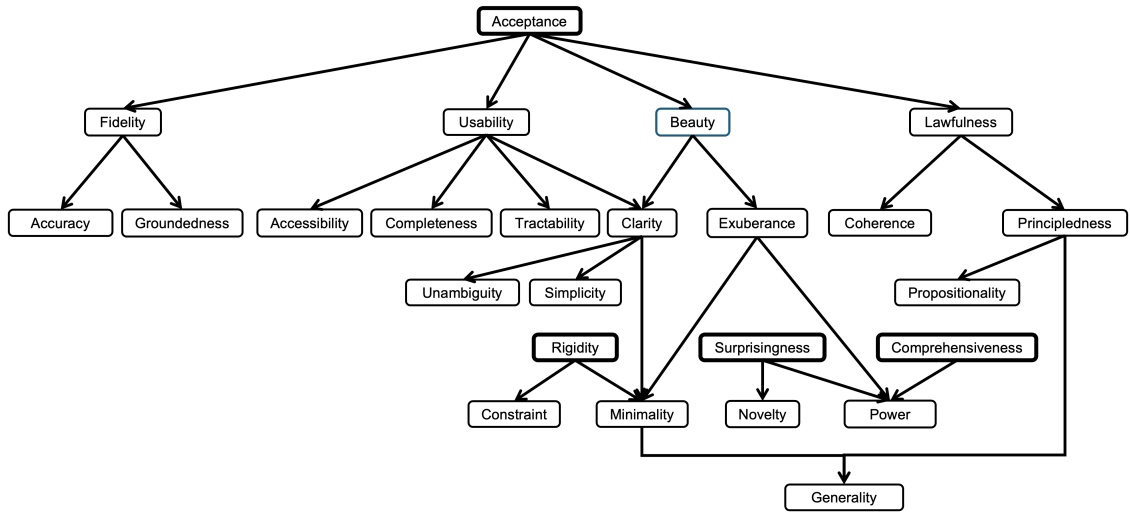

Figure 1: Graph of evaluation criteria for theories. An arrow from one criterion to another indicates that the former depends on the latter. The roots of the graph, highlighted in bold, have no other criteria depending on them. Adapted from Rosenbloom (2026).

Relevant prior work on theory evaluation can also be found more broadly in philosophy of science (e.g., Keas, 2018). But the point here is not to argue that this graph necessarily provides the best such set of criteria – it is quite possible to consider additional criteria as well as simplifications and reorganizations of the criteria already listed here – rather that the graph, as is, enables an informative, albeit qualitative at this point, comparative analysis of the criterial advantages and disadvantages of cognitive versus generative theories that extends significantly beyond simply which theories provide more accurate models of the phenomena they are about.

This comparison proceeds here across two levels. The first level considers just the architectures themselves as theories. The second level considers the full systems that result when variable content is included. While architectural comparison may be considered to be at the heart of this work, some criteria can be difficult to evaluate just based on architectures – instead requiring full systems to do them complete justice – while the direction of other criteria may actually flip when variable content

---

in technology are in many ways more in the tradition of cognitive architectures than of generative architectures.

and the resulting full systems are considered. Thus, a complete picture requires considering both architectures and systems.

The specific criteria considered in this work are focused on two roots of the overall graph in Figure 1 – *acceptance* and *comprehensiveness* – plus all of the other criteria on which they depend, whether directly or indirectly. Acceptance is the primary root of the graph. It straightforwardly concerns whether the theory is accepted by the relevant scientific community. This is the ultimate criterion for any theory to have an impact moving forward, and it depends directly or indirectly on nearly all of the other criteria in the graph. Acceptance depends directly on four major criteria to be considered here: *fidelity*, *lawfulness*, *usability*, and *beauty*. These four criteria turn out to map roughly onto the *semantics*, *syntax*, *pragmatics*, and *aesthetics* of a theory, thus effectively treating theories as linguistic expressions that are to be evaluated via corresponding criteria.

Not all of these criteria will be relevant to all theories or all theorists, depending on both the intended uses of the theories and the research strategies of the theorists, but all are potentially relevant to theories of whole minds. Comprehensiveness, although a more minor root, is included here because it is particularly relevant for theories of whole minds. It thus provides a fifth major criterion to be considered. The graph in Figure 1 also includes two additional minor roots – *rigidity* and *surprisingness* – but they are not considered further here.

The following five sections discuss the corresponding criteria along with the comparative advantages of cognitive and generative theories with respect to them. Sometimes this extends to actual comparative evaluations but at other times it is limited to just the comparative feasibility of performing such evaluations. The results reveal a rich space of tradeoffs that go beyond simply how well systems of the two types may perform on benchmarks. The final section concludes with a summary of what has been learned.

## 2. Fidelity

Fidelity (Figure 2) tends to be the dominant criterion across the sciences, concerning primarily a theory's overall *accuracy* with respect to the phenomena of interest. Recursively, it also concerns the level of support not just for the theory as a whole but also for the individual pieces out of which it is constructed, via a criterion that can be termed *groundedness*.

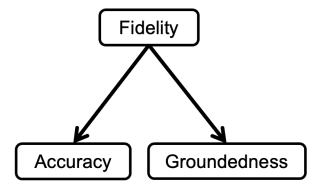

Figure 2: Dependencies for *Fidelity*. Adapted from Rosenbloom (2026).

### 2.1 Architectural Fidelity

Behavioral fidelity is difficult to assess for architectures in isolation, as without variable content there is typically little to no substantive behavior to compare against phenomena of interest, except for perhaps early-stage learning behavior. Still, partial assessments of accuracy or groundedness may be possible. One simple example from the physical sciences is Newton's law of universal gravitation: $F=G(m_1 \times m_2)/r^2$. The architecture – that is, the law's fixed structure – here is simply the equation, with the variable content being the masses, the distance, and possibly the gravitational constant. Without values for any of these parameters it is still possible to accurately predict that larger masses and shorter distances yield increased gravitational force.

In the world of mental architectures – particularly for cognitive architectures – something close to an architectural evaluation may be possible when a minimum of variable content is required for behavior, such as when the phenomena of interest occur at very short time scales where it is largely the architecture that shows through. Evaluation of structural accuracy may also possible; for example, by comparing the components out of which transformers are built with structures in the human brain (e.g., Kozachkov, Kastanenka & Krotov, 2023), or by comparing the presence of procedural and declarative memories in cognitive architectures with the overall structure of human memory (e.g., Squire, 1987). A more detailed cognitive example can be found in the evaluation of the Common Model of Cognition – a community consensus concerning what must be in an architecture for humanlike cognition (Laird, Lebiere & Rosenbloom, 2017) – as a high-level architecture for the human brain (Stocco *et al*., 2021). With no variable content, and in fact without even an implementation of the model, predictions from the Common Model concerning connectivity among functional circuits in the human brain turned out to be more accurate than standard models from neuroscience.

**2.2  System Fidelity**

Behavioral fidelity becomes much easier to evaluate when applied to systems as a whole, as the behavior of such systems can be compared directly to that of either human subjects or abstract benchmarks. For cognitive architectures, this typically amounts to building models of behavior – or phenomena – based on the architecture plus limited amounts of variable content, whether in the form of procedural skills or declarative knowledge. When these models only tickle a small fraction of the architecture's full capabilities – as is often the case – the comparisons may bear more directly on grounding than accuracy, but either way such work can contribute significantly to assessing fidelity (modulo concerns about rigidity – a criterial root alluded to in the introduction – that may arise from the flexibility of the variable content).

Still, evaluating cognitive systems as a whole for accuracy across a broad range of types of behaviors, rather than as a collection of models developed and tested individually, has also proven challenging. There are exceptions, such as the development of *cognitive supermodels* that use a single system "to account for behavior across a range of diverse domains" (Salvucci, 2010); and work on *interactive task acquisition*, in which a single system learns how to perform a diversity of tasks through human instruction in the context of attempts at performing the tasks (Gluck & Laird, 2018). But such assessments are difficult in general due to the lack of sufficient bodies of knowledge and skills. Architectural incompleteness (Section 4.2) and difficulties with interactions among the mechanisms actually embodied can also make this even more challenging.

Generative systems tend not to suffer from these issues, as they are trained on vast amounts of data – yielding significant *power* in terms of the breadth of phenomena covered (Section 5) – and the training ensures that the interactions among their small number of basic mechanisms work well at least for achieving accuracy in predicting the next word. They thus can be evaluated for accuracy across quite broad ranges of desired behavior. However, given that they are trained explicitly just for this purpose – at least in their purest form – their accuracy can suffer greatly in this broader context, in the form of what are known as *hallucinations*.

Some whole-mind theories have traditionally sacrificed some amount of fidelity for power; that is, the breadth of phenomena covered (Section 6). That has been seen, for example, in Soar (Newell, 1990) on the cognitive side in any of the many generative theories that hallucinate extensively. This may be justified simply by the application benefits of power or by the notion that if the ultimate goal is to achieve both power and fidelity it may be easier to incrementally improve fidelity once power has been maximized than to increase power once fidelity has been maximized. In other terms, as with abstraction planning (Sacerdoti, 1974), starting with an approximate global model before attempting to fill in the details may work better than attempting to extend one or more local optimizations.

While not necessarily explicitly so, work on post-training in generative systems turns out to fit this abstraction-planning model quite well. Consider for example, Centaur (Binz *et al*., 2024), a form of cognitive supermodel that takes as its starting point a version of the Llama 3 large language model (LLM) (Grattafiori, 2024) and then specializes it via additional training over the results of 160 psychological experiments. Such an approach can ameliorate the tradeoff between fidelity and power, although Centaur itself – as with all generative theories – still implicates other tradeoffs, with respect to criteria such as lawfulness (Section 3) and usability (Section 4). On the cognitive side, my own view of Soar has long been based on such an analogy. Yet, other cognitive theories – particularly ones like ACT-R (Anderson *et al*., 2004) that strongly prioritize cognitive science over AI or AGI – may subordinate power to fidelity from the very beginning. None of these theories, whether cognitive or generative, has however been terribly good at yielding single systems that simultaneously achieve high fidelity across a broad range of domains.

## 2.3 Fidelity Summary

Architectures of both types are difficult to evaluate for fidelity, although there are exceptions when the time scales are small (for cognitive architectures) or the comparison is of structural aspects. In contrast, systems of both types can typically be evaluated for fidelity. If there is a tradeoff of some amount of fidelity for power, this may be overcome via an analog of abstraction planning for particular domains, but it is not typical to see many such high-fidelity domains combined into a single system. Finally, systems with comparable power – as is typical of generative systems – are more amenable to comparison via standardized benchmarks.

## 3. Lawfulness

Lawfulness (Figure 3) concerns the form of the theory – in terms of its *coherence* and the extent to which it is expressed in terms of *principles* – independent of its relationship to the phenomena of concern. These principles are typically assumed to take the form of *propositions* that individually exhibit *generality*. Lawfulness is sometimes considered definitional for a theory (e.g., "Theory," 2024), but it can alternatively be viewed as comprising one or more dimensions along which theories may vary, and thus be applied as a means of evaluating them. In a sense, lawfulness can be seen as preferring neatness in the old neat versus scruffy debate in AI.

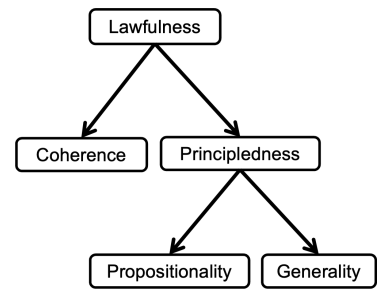

Figure 3: Dependencies for *Lawfulness*. Adapted from Rosenbloom (2026).

### 3.1 Architectural Lawfulness

Both cognitive and generative architectures typically rate high on coherence, given that they are the result of careful design processes. However, architectures that can be cast in either a mathematical or logical form may have an advantage over more procedurally defined architectures in being amenable to proofs of coherence. This spans all generative architectures and, in part or whole, some cognitive architectures.

Principledness decomposes into propositionality and generality, where the latter refers to the generality of the individual parts of the theory rather than of the theory as a whole. Both types of architectures typically support generality. Where they most often differ is in propositionality. Generative architectures typically have a large advantage here. Even if they are actually implemented in a procedural fashion – that is, as code – the ability to express them in terms of a relatively small number of general mathematical equations yields a strong form of propositionality. Cognitive architectures can, however, approach this when they can be specified in logical or mathematical terms (e.g., Milnes *et al*., 1992; Cooper *et al*., 1996; Hutter, 2005).

### 3.2 System Lawfulness

Cognitive systems typically have a strong advantage over generative systems with respect to both aspects of lawfulness due to the symbolic nature of much of their variable content. Employing symbolic representations does not guarantee lawfulness, but it does provide a better basis for it than masses of parameter values. This is one of key shortfalls of Centaur alluded to in Section 2.2.

Much of the variable content in cognitive architectures is also hand coded, which can encourage lawfulness while simultaneously reflecting lack of *completeness* (Section 4.2) in their learning ability. With generative architectures, only the links are typically hand coded, yielding a marginal amount of this "benefit." Their parameter values are learned, and yield a body of variable content for which both aspects of lawfulness only exist, when they exist at all, as a side effect of optimizing according to the architecture's learning criteria. This lack of lawfulness is likely a significant contribution to their production of hallucinations. Attempts at explaining generative systems (Zhao *et al*., 2024) can be seen as one approach to ameliorating their inherent lawlessness but, even when this possible, it at best provides a cognitive theory of the generative theory.

### 3.3 Lawfulness Summary

There is a peculiar tradeoff here with respect to these two classes of theories. There is a propositionality advantage of generative architectures over procedurally defined cognitive architectures that is part of their overall appeal with respect to cognitive architectures, but the relative lack of lawfulness in generative systems is a key factor in both their production of hallucinations and the difficulty in explaining them, and thus in the lack of trust that can result.

## 4. Usability

Usability concerns the relationship of theories to those who use them. This is typically in terms of human users but also must include the computers that execute implementations of them. Almost tautologically an unusable theory is unuseful, and thus unlikely to achieve acceptance by the scientific, or any other, community. As shown in Figure 4, usability directly depends on four subcriteria: *accessibility*, *completeness*, *tractability*, and *clarity*. In contrast to the approach taken in the other sections of this paper, of first covering architectures and then systems, due to the number of subcriteria here the structure is inverted to discuss both architectures and systems for each top-level subcriterion in a single subsection.

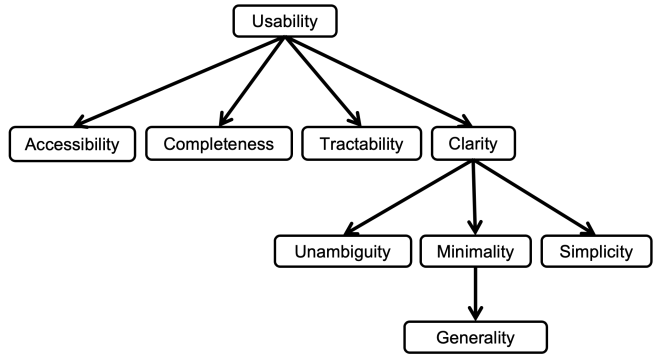

Figure 4: Dependencies for *Usability*. Adapted from Rosenbloom (2026).

### 4.1 Accessibility

Accessibility concerns whether the theory is articulated in a medium accessible across individuals, rather than simply being in the mind of a single individual. Like lawfulness, it may be considered definitional, as in Newell's (1990) use of *explicitness*, but it can also correspondingly be considered as an evaluation criterion that affects the usability of the theory by a community of users. Still, it will be assumed for all of the theories considered here, and thus not discussed further.

### 4.2 Completeness

Completeness concerns whether the theory includes all of the parts necessary to answer the questions it is to cover. If there are many pertinent questions, partial completeness can lead to the ability to answer some but not all of them. With respect to architectures, completeness focuses on mechanisms. Both cognitive and generative architectures typically have System 1 (Kahneman, 2011) performance mechanisms – or what Newell (1990) referred to as *knowledge search* – effectively yielding knowledge-driven reactivity. But generative architectures notably lack, at least, online learning and the kinds of System 2 capabilities – or what Newell (1990) referred to as *problem space search*, but with a particular focus here on metacognitive and reflective abilities – found in many cognitive architectures. Since these gaps may lead to incorrect behavior rather than a failure to behave, they may be considered as issues with fidelity rather than completeness, but it seems cleanest to categorize them as architectural incompleteness rather than system infidelity.

With respect to systems, they can be considered complete if they can answer the questions of interest. Here, generative theories shine due to the vast amounts of data on which they are trained. There is no reason in principle that cognitive architectures cannot achieve the same level of completeness – unless tractability (Section 4.3) prevents this – but they do not do so at present unless they incorporate something like a generative AI component (e.g., Romero *et al*., 2023). Completeness can rather trivially be achieved at the expense of power simply by restricting the questions of interest to those the system can actually answer, but that is rather a hollow approach.

### 4.3 Tractability

Tractability for computational theories concerns whether it is feasible to compute answers to the questions of interest. This is not limited to the traditional computer science notion of a computation being subexponential, including on either side of this whether the problem simply doesn't take too long to solve or is even computable. Tractability typically concerns interactions between the architecture and variable content, with distinct issues showing up for System 1 versus System 2 aspects of the architecture. It can also be split orthogonally according to combinatoric versus real-time tractability, with the former focusing on how time grows with the growth of the variable content and the latter focusing on whether processing completes in sufficient time for the task being pursued.

As discussed in the previous section, pure generative architectures only implicate System 1. Neither their performance nor learning algorithms are combinatoric computationally, so the core question for both concerns real-time behavior. Given that large LLMs may learn trillions of parameters from petabytes of data, this is a significant issue that has led to much research on how to improve the situation (Wan *et al*., 2023); and on a smaller scale, it may have also contributed to the earlier years-long delays in the development and application of neural networks in general. Part of the current real-time issue may, of course, also reflect the need to learn from a combinatoric set of instances to counterbalance an inability to deal with such combinatorics internally.

Cognitive architectures also implicate System 1 for both performance and learning. They typically do not, however, have the same kind of real-time issue for learning, as they acquire new knowledge incrementally and online. They can often also handle combinatoric possibilities internally via System 2. With respect to performance, they have typically dealt with much smaller memories; but, even so, work has been required on handling larger procedural (Minton, 1988; Doorenbos, 1993) and declarative (Douglass, Ball & Rodgers, 2009; Derbinsky, Laird & Smith, 2010) memories.

As implied above, cognitive architectures may introduce combinatorics not found in generative architectures. For System 1, this may for example result from individual elements of procedural memory becoming intractable to use (Tambe, Newell & Rosenbloom, 1990). For System 2, combinatorics typically shows up when cognitive architectures support problem space search (Newell *et al*., 1991; Laird, 2012; Rosenbloom, Demski & Ustun, 2016a). They typically attempt to cope with this via the addition of metalevel control knowledge, whether programmed or learned, that limits the size of the search (Laird, Newell & Rosenbloom, 1987).

### 4.4 Clarity

Clarity depends on a combination of three further criteria: *unambiguity*, *minimality*, and *simplicity*. Unambiguity concerns whether there is a clear single interpretation, versus multiple plausible ones. Minimality concerns the size of the theory, assuming smaller is better, in rough alignment with Occam's razor. As with principledness, minimality also depends on the generality of the elements of the theory (Section 3), given that smaller theories for the same body of phenomena must inherently be built from more general parts. Simplicity concerns how easy the theory is to interpret, independent of its size or degree of ambiguity. For example, complex mathematical expressions may be small and unambiguous while sacrificing simplicity for some users in this sense.

Architectural designs on paper, such as The Society of Mind (Minsky, 1986) and the Common Model of Cognition (Laird, Lebiere & Rosenbloom, 2017), may easily introduce issues of ambiguity. One of the reasons that implementation of architectures is so important in general is that it provides one of the major means of ensuring the absence of ambiguity. Thus, implemented architectures of all types are typically free of this form of ambiguity.

With respect to minimality, developers of architectures may hold a physics mindset that particularly emphasizes it in the architecture (e.g., Hutter, 2005; Rosenbloom, Demski & Ustun, 2016a; Vaswani *et al.*, 2017; Silver *et al.*, 2021); or a biological mindset that is much more comfortable with an efflorescence of mechanisms (e.g., Goertzel, Pennachin & Geisweiller, 2014); or a more neutral mindset that while preferring minimality is more amenable to trading it off for other considerations (e.g., Anderson *et al.*, 2004; Laird, 2012). Generative architectures are typically more minimal than cognitive architectures, but it is unclear how much of this is due to their relative incompleteness with respect to cognitive architectures.

With respect to simplicity, generative architectures are simpler than cognitive architectures, with the former typically amounting to sets of equations that should be clear to anyone with the appropriate mathematical background and the latter typically involving either a complex description in natural language or a complex body of code. For discriminative neural networks, the former has enabled even high school students to contribute papers at leading conferences (Lee, 2025). Still, the relative simplicity of generative architectures may at least part be due to their greater degree of incompleteness.

When it comes to systems, generative architectures can suffer significantly compared to cognitive architectures in terms of both unambiguity and simplicity. This is the other key shortfall of Centaur alluded to in Section 2.2. Minimality is less clear without a better understanding of the compression yielded by backpropagation in generative systems versus what is provided by general symbolic structures.

**4.5 Usability Summary**

Generative architectures are typically more tractable, simpler, and more minimal than cognitive architectures whereas cognitive architectures are typically more complete. Cognitive systems are typically more tractable, simpler, and more unambiguous than generative systems, whereas generative systems are typically more complete than cognitive ones. However, the story on tractability is really much messier than these comparisons would seem to imply. In particular, further completing generative architectures and the variable content in cognitive systems could change these assessments dramatically.

**5. Beauty**

Beauty is inherently aesthetic, but it is also assumed to have heuristic value, at least in parts of physics (Wilczek, 2015), in leading towards theories that are more likely to be accurate in modeling natural phenomena. As shown in Figure 5, it depends on a combination of *clarity* (Section 4.4) and *exuberance*. Exuberance amounts to a relationship between *minimality* (Section 4.4) and *power*, or in every-day terms how much bang for the buck the theory yields.

As mentioned in Section 2.2, power concerns the breadth of phenomena covered by the theory. It is related to *completeness* (Section 4.2), but rather than focusing on (lack of) gaps with respect to questions to be answered it focuses on the breadth of what is possible.

In the study of the mind this would typically be viewed as generality rather than power, but power is a common term more broadly in science, and generality has already been used in a different way in the context of principledness – concerning the individual elements of the theory rather than the theory as a whole. No matter what name it goes by, in the form of *comprehensiveness* (Section 6) power is largely the raison d'être of theories of whole minds. Thus, irrespective of whether there is a concern for beauty, or for its direct dependence on clarity and exuberance, or even for their joint dependence on minimality, this aspect of beauty is critical.

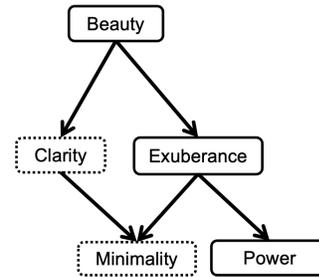

Figure 5: Dependencies for *Beauty*. Dotted outlines are for criteria whose dependencies are shown in Figure 4. Reproduced from Rosenbloom (2026).

Because of the dependence of exuberance on minimality, it too can be seen to align with Occam's razor – essentially, preferring the smallest theory for a given amount of power. Exuberance also can be seen to align with approaches concerned with the level of compression embodied by a theory (Wolff, 2013). Other particular notions, such as the *free energy principle* (Friston, 2010), or that *reward is enough* (Silver *et al*., 2021), also clearly embrace exuberance.

In being partly based on clarity, beauty's definition here may overlap with how it is used in the arts, but it can also diverge in significant ways – ambiguity, size, and complexity can all conceivably play significant roles in the beauty of particular artistic creations. Exuberance is also a term that can be relevant in the arts, as for example found in the baroque period that spanned parts of the 17th and 18th centuries. But the meaning here specifically reflects how much phenomena – that is, how much power – can be spanned by how little theory. Power too is a term found in the arts, but there it instead concerns the impact on the audience, or "user."

## 5.1 Architectural Beauty

Starting at the bottom of Figure 5, with power, the most natural measure of power for an architecture concerns the range of intelligent capabilities – such as reasoning, problem solving, planning, learning, and natural language processing – it can exhibit. Here, cognitive architectures have a significant edge over generative architectures, being typically richer in, at least, memory structures, learning mechanisms, and reasoning and metareasoning capabilities. But only time will tell whether this is simply a sign of the immaturity of generative architectures versus a more inherent limitation.

As discussed in Section 4.4, minimality is more typical of generative than of cognitive architectures, although some cognitive architectures do strive for this. In combination with the just mentioned concern about the lack of power of existing generative architectures, this yields only a qualified advantage with respect to minimality for generative architectures.

Combining power and minimality, to yield exuberance, shifts the focus to how many mental capabilities can result from combinations of how few architectural mechanisms. This is not a typical concern in cognitive architectures, although it has been important, for example, in both Sigma (Rosenbloom, Demski, & Ustun, 2016a) and the early versions of Soar (Laird, Newell & Rosenbloom, 1987). This is quite explicit in Sigma, with its desideratum of *functional elegance*. Typically, such exuberance actually requires combining architectural mechanisms plus at least small amounts of variable content – as illustrated for example in the early work on a *universal weak method* in Soar (Laird & Newell, 1983) – rather than solely being due to combinations of architectural mechanisms. Still, most of the action is due to the architectural mechanisms.

With the development of transformers, the notion that *attention is all you need* (Vaswani *et al.*, 2017) is an exuberant claim that is now central to generative architectures. Experience with generative systems has revealed a range of intelligent capabilities apparently exhibited by them, yet how much of this should be allocated to architectural rather than system exuberance – that is to large quantities of variable content – remains to be seen, with only a murky answer thus possible at present as to which class of architectures is more exuberant.

Moving further up the hierarchy, to beauty, the most beautiful architectures are both clear and exuberant, favoring those that are expressed mathematically or logically in a manner that enables much to come out of little. Such theories can be found, for example, in physics, but it remains an open question as to whether such an idealization is possible for minds, as well as which approaches might ultimately yield the most beautiful theories. At present, cognitive architectures tend to exhibit power at the expense of clarity whereas generative architectures tend to exhibit clarity at the expense of power.

One class of approaches that spans paradigms and yet may be particularly promising for beauty combines some form of induction with reinforcement learning, whether in the form of an AGI architecture such as *Universal AI* (Hutter, 2005), or a discriminative neural network architecture such as deep reinforcement learning (Francois-Lavet *et al.*, 2018), or a generative architecture such as large language models with reinforcement learning from human feedback (Ziegler *et al.*, 2019). These all start with high levels of clarity but must further establish their exuberance. Sigma's *graphical architecture*, which implements its cognitive architecture via an extended form of factor graphs (Kschischang, Frey & Loeliger, 2001), combines significant power – including directly supporting induction and indirectly supporting reinforcement learning when augmented with appropriate variable content – with an approximation of the mathematical ideal for clarity (Rosenbloom, Demski & Ustun, 2016b).

### 5.2 System Beauty

When the focus is shifted to entire systems, it makes sense to measure power in terms of the range of problems, or whole problem domains, over which intelligent behavior can be exhibited. Although cognitive systems have been applied across a wide variety of domains, the breadth of training available to generative systems implies that, at least at present, any single generative system has a huge power advantage over any individual cognitive system. In the other direction, although there is no clear advantage with respect to minimality (Section 4.4), there is a clear advantage for cognitive systems in terms of clarity (Section 4.4). All systems thus currently lack either power or clarity and therefore do not come close to overall beauty. Still, cognitive systems at least provide a possible path, based on extending their variable content, whereas even explainable generative systems would still fall short of overall beauty.

### 5.3 Beauty Summary

There is a particularly complex combination of tradeoffs between cognitive and generative theories with respect to beauty. Cognitive architectures typically yield more power whereas generative architectures typically yield more clarity. Yet, conversely, cognitive systems typically yield more clarity whereas generative systems typically yield more power. Aa result, neither class of theories can currently be said to yield either beautiful architectures or systems, although the ultimate possibility of either cannot yet be ruled out.

## 6. Comprehensiveness

Comprehensiveness concerns whether the range of phenomena covered by a theory is exhaustive with respect to some natural maximal domain. When physicists talk about a "Theory of Everything*"* (2024), they have in mind a comprehensive theory of the universe. More on point, unified theories of cognition (Newell, 1990) and AGI (Goertzel, 2014) strive for comprehensive theories of human or artificial intelligence. Comprehensive mental architectures contain all of the mechanisms required to yield general intelligence, whether the goal is human(like) or artificial intelligence. Comprehensive systems combine comprehensive architectures with all of the variable content required to perform intelligently across, at least, the domains of human performance. As shown in Figure 6, comprehensiveness depends on power, but it also depends on this notion of what the maximal domain is that isn't itself considered a criterion here.

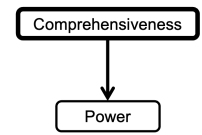

Figure 6: Dependency for *Comprehensiveness*. Adapted from Rosenbloom (2026).

### 6.1 Architectural Comprehensiveness

Both cognitive and generative architectures fall short with respect to comprehensiveness, although in different ways. As mentioned in Section 4.2, generative architectures fall significantly short of what is required, particularly with respect to such things as online learning and System 2 capabilities. However, the inclusion of reinforcement learning from human feedback (Ziegler *et al*., 2019) or chain-of-thought prompting (Wei *et al*., 2022) does add limited forms of the latter. More ambitious attempts to build broader architectures around generative architectures are also in progress (e.g., Park *et al*., 2023; Sumers *et al*., 2023).

Cognitive architectures are typically more comprehensive in this sense than generative architectures, including significant System 1 and 2 capabilities. However, they still do fall short, with each architecture tending to fall short in its own unique way. This is a key reason, at least in my view, why there has been essentially no success over the years in developing useful benchmarks for progress with cognitive architectures.

### 6.2 System Comprehensiveness

Generative systems tend to be the most comprehensive due to the vast amounts and varieties of data from which they learn. This is also likely why benchmarks have found a role in evaluating

EVALUATION OF COGNITIVE AND GENERATIVE THEORIES

such systems (e.g., Evidently AI Team, 2024), as it can be assumed that they are all comprehensive enough to be compared with each other via the same benchmarks.

Cognitive systems that leverage knowledge graphs (Rytting, 2000; Forbus & Hinrichs, 2006; Douglass, Ball & Rodgers, 2009) reach partway towards this, with ones that incorporate generative AI's as modules (e.g., Romero *et al*., 2023) approaching even closer. Otherwise, cognitive systems tend actually to only span limited classes of tasks due to the minimal amount of variable content, in the form of knowledge and skills, that they contain.

**6.3 Comprehensiveness Summary**

For comprehensiveness the tradeoff between cognitive and generative architectures is rather straightforward at this time, with cognitive architectures being more comprehensive than generative architectures but generative systems being more comprehensive than cognitive systems. Combining the two approaches yields one potential path towards completeness at both levels.

**7. Conclusion**

Considering theory evaluation in terms of fidelity, lawfulness, usability, beauty, and comprehensiveness enables a broader, albeit qualitative, understanding of the strengths, weaknesses, and tradeoffs between cognitive and generative theories.

Bare architectures of any sort can be tough to evaluate for fidelity, although some forms of evaluation are possible. Systems of either type can be evaluated, albeit with generative ones being better at enabling the use of shared benchmarks. Still, the drive for power – and even for comprehensiveness – can lead to sacrificing some amount of fidelity, at least in the short term.

With respect to both lawfulness and usability, generative theories have the overall advantage with respect to the architectures themselves, whereas cognitive theories have the advantage with respect to the variable content processed by the architectures. With respect to beauty, there is a complex set of tradeoffs between the two types of theories, with an inversion of criteria occurring when comparing architectures versus systems. Overall, neither type of theory is able to yield truly beautiful systems at this point. With respect to comprehensiveness, there is currently a simple tradeoff, with cognitive architectures and generative systems each having the advantage.

Future work on such comparative evaluations is possible along multiple paths, including: (1) refining the set of criteria; (2) applying the full set of criteria in Figure 1 rather than just a subset of them; (3) extending the comparisons from qualitative to quantitative; (4) broadening the classes of theories considered to include whole brain theories and narrower theories that don't strive to cover whole minds or brains; and (5) comparing subclasses of, and even individual, theories. Future work also makes sense in the direction of developing theories that perform well across this broad range of criteria, whether by integrating across these two classes of theories or via other approaches.